\documentclass[11pt]{article}
\usepackage[utf8]{inputenc}
\usepackage[australian]{babel}

\usepackage[T1]{fontenc}
\usepackage{lmodern} 
\usepackage{microtype}

\usepackage{xcolor}
\definecolor{brandonblue}{RGB}{0,0,192}  
\definecolor{brandondarkgrey}{RGB}{25,25,25}  

\usepackage{hyperref}

\usepackage{setspace}
\usepackage[
  a4paper,
  hmargin=1in,
  vmargin=1in,
  includeheadfoot, 
  nohead,
]{geometry}


\usepackage{titlesec} 
\titleformat{\section}[block]{\normalfont\bfseries\Large}{\thesection.}{0.5em}{}
\titlespacing*{\section}{0em}{1ex}{-1ex}
\titlespacing*{\subsection}{0em}{1ex}{-1ex}

\usepackage{fancyhdr}

\usepackage[lining]{nunito} 

\hypersetup{pdfpagemode={UseNone},pdfstartview={FitV},pdfpagelayout={SinglePage},unicode,bookmarksopen=false,colorlinks,allcolors={brandonblue},breaklinks=true} 
\usepackage{amsmath,amssymb,amsfonts} 
\usepackage{tikz} 

\usepackage[math]{cellspace} 
\usepackage{graphicx}
\usepackage{placeins} 

\usepackage[
  justification=centerlast,
  font={small,stretch=1.3},
  labelfont={bf},
]{caption}
\usepackage{enumitem} 
\setlist{topsep=0pt,beginpenalty=10000,first=\interlinepenalty10000}

\usepackage{csquotes} 
\usepackage[multiple]{footmisc} 

\usepackage[square,numbers,sort,nonamebreak]{natbib}

\newcommand{\figref}[1]{Figure~\ref{#1}}

\newcommand{\secref}[1]{Section~\ref{#1}}

\definecolor{mydarkred}{RGB}{136,0,23}
\definecolor{mymedred}{RGB}{211,144,155}
\definecolor{mylightred}{RGB}{235,204,209}
\newcommand{\ColouredCircleK}{\protect\tikz\protect\draw[draw=none,fill=black] (0,0) rectangle (1.5ex,1.5ex) ;}%
\newcommand{\ColouredCircleD}{\protect\tikz\protect\draw[draw=none,fill=mydarkred] (0,0) rectangle (1.5ex,1.5ex) ;}%
\newcommand{\ColouredCircleM}{\protect\tikz\protect\draw[draw=none,fill=mymedred] (0,0) rectangle (1.5ex,1.5ex) ;}%
\newcommand{\ColouredCircleL}{\protect\tikz\protect\draw[draw=none,fill=mylightred] (0,0) rectangle (1.5ex,1.5ex) ;}%

\begin{document}
\pagestyle{fancy}
\fancyhf{}
\renewcommand{\headrulewidth}{0pt}
\fancyfoot[R]{\color{brandondarkgrey} \footnotesize \thepage}

\begin{center}
    {\LARGE ServoLNN: Lagrangian Neural Networks Driven by\\[1ex] Servomechanisms}\\[0.5\baselineskip]
    Brandon Johns\footnotemark[1] \href{https://orcid.org/0000-0002-8761-5432}{\includegraphics[height=2ex]{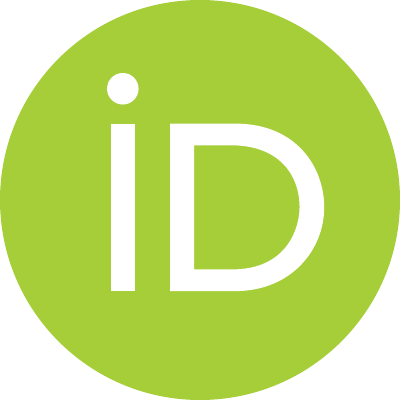}},
    Zhuomin Zhou\footnotemark[1] \href{https://orcid.org/0009-0005-7268-7256}{\includegraphics[height=2ex]{ORCIDiD_iconvector.pdf}},
    Elahe Abdi\footnotemark[1] \href{https://orcid.org/0000-0003-3748-0442}{\includegraphics[height=2ex]{ORCIDiD_iconvector.pdf}}\\[0.3\baselineskip]
    \footnotemark[1]Department of Mechanical and Aerospace Engineering, Monash University, Victoria, Australia\\
    \href{mailto:brandon.johns@monash.edu}{brandon.johns@monash.edu},
    \href{mailto:zhuomin.zhou@monash.edu}{zhuomin.zhou@monash.edu},
    \href{mailto:elahe.abdi@monash.edu}{elahe.abdi@monash.edu}\\
    \rule{\textwidth}{1.6pt}\vspace*{-1\baselineskip}\vspace*{2pt} 
    \rule{\textwidth}{0.4pt}\\[0.5\baselineskip] 
\end{center}

\vspace*{-1\baselineskip}{ \footnotesize This document is a preprint. February 2025. }

\vspace*{-0.8\baselineskip}
\begin{displayquote}
\textbf{\textit{Abstract} \textemdash}
Combining deep learning with classical physics facilitates the efficient creation of accurate dynamical models. In a recent class of neural network, Lagrangian mechanics is hard-coded into the architecture, and training the network learns the given system. However, the current architectures do not facilitate the modelling of dynamical systems that are driven by servomechanisms (e.g. servomotors, stepper motors, current sources, volumetric pumps).

This article presents ServoLNN, a new architecture to model dynamical systems that are driven by servomechanisms. ServoLNN is compatible for use in real-time applications, where the driving motion is known only just-in-time. A PyTorch implementation of ServoLNN is provided.

The derivations and results reveal the occurrence of a possible family of solutions that the training may converge on. The effect of the family of solutions on the predicted physical quantities is explored, as is the resolution to reduce the family of solutions to a single solution.

Resultantly, the architecture can simultaneously accurately find the energies, power, rate of work, mass matrix, generalised accelerations, generalised forces, and the generalised forces that drive the servomechanisms.
\end{displayquote}
\vspace*{-1\baselineskip}
\rule{\textwidth}{0.4pt}\vspace*{-1\baselineskip}\vspace{3.2pt} 
\rule{\textwidth}{1.6pt}

\section{Introduction} \label{sec:intro}
Accurately modelling the dynamics of a given physical system can be a challenging task. Commonly, establishing system equations requires applying a set of simplifying assumptions. However, the appropriate choice of assumptions can be difficult to determine, and the behaviour of the model can significantly vary between choices \cite{BJ-04}. Likewise, the appropriate choice of numerical values for the system parameters is not always clear \cite{BJ-04}.

Deep learning can aid in the creation of accurate dynamical models. It allows embedding assumptions within the model or loss function, or making no assumptions and only learn from data \cite{RF1-06}. For example, learning could be used to obtain the parameters for an analytically established model, or learning could be used to fit a `black box' model that does not embed any prior knowledge about the given system or the laws of physics \cite{RF1-06,RF1-05}. However, the former approach does not address the problem of the designer potentially embedding inappropriate assumptions into the analytical model, and the latter approach is inefficient in having to learn the well-established principles of classical physics.

This leads to two distinctions. Firstly, knowledge about the laws of physics can be treated differently from knowledge about the given system. Secondly, physics-guided architecture is distinct from physics-guided loss \cite{RF1-05}. Physics-guided architecture hard-codes mathematical relations into the architecture (as opposed to using a black box model) in a manner that intrinsically constrains the output to obey these relations \cite{RF1-05}. Physics-guided loss is less rigorous in only penalising the breaking of mathematical relations during training \cite{RF1-05}. Notably, penalising the breaking of physical laws does not prevent them from being broken, but hard-coding them into the architecture does.

A new class of neural network architectures has recently emerged, stemming from Deep Lagrangian Networks (DeLaN) in 2019 \cite{RF1-04}, which hard-codes classical physics into the architecture while making minimal assumptions about the given system. Lagrangian or Hamiltonian mechanics is hard-coded into the architecture, and training the network learns the given system. This physics-guided architecture can also be coupled with physics-guided loss to impose additional known constraints on the system \cite{RF1-08}.


This class of neural network is versatile in that it yields many different physical quantities from just one evaluation of the network, and these quantities are guaranteed to be consistent with each other \cite{RF1-06}. This permits analysis of the internal state of the dynamical system. The quantities that can be found include the energies, power, rate of work, mass matrix, generalised accelerations, generalised forces, and some decompositions thereof \cite{RF1-06,RF1-03}. It is notable that a single trained network of this architecture can evaluate both forward dynamics and inverse dynamics \cite{RF1-06}.

A recent review of DeLaN-derived architectures is \cite{RF1-06}, and a review on the broader topic of neural networks for simulating dynamical systems is \cite{RF1-23}. A review on the topic of physics-guided deep learning algorithms is \cite{RF1-18}. The topic is also discussed in a review in the context of structural dynamics and vibroacoustics \cite{RF1-05}. These reviews class the DeLaN-derived architectures as a neural network to represent the dynamical model. This is contrasted to other classes of neural network that only represent specific solution trajectories. Thus, a trained network with a DeLaN-derived architecture is reusable with different initial conditions and inputs \cite{RF1-23}. The incorporation of physics principles additionally specialises the DeLaN-derived architectures compared to the general architectures to represent ordinary differential equations (ODEs) \cite{RF1-23,RF1-18}.

Some notable DeLaN-derived architectures are discussed as follows. DeLaN \cite{RF1-04} separately outputs the kinetic and potential energies. The architecture intrinsically guarantees the positive definiteness of the mass matrix. External forces are permitted to act on the system. DeLaN 4EC \cite{RF1-03} extends this with the incorporation of a friction model, and a physics-guided loss function that enforces the first law of thermodynamics and differential smoothness of the energy functions. SymODEN \cite{RF1-20} and Dissipative SymODEN \cite{RF1-21} are similar to DeLaN, but implement Hamiltonian mechanics instead of Lagrangian mechanics. A function that relates the control input to the generalised force is also included in the network architecture. CHNN \cite{RF1-08} incorporates Lagrange multipliers with user-specified holonomic constraint equations to improve accuracy and efficiency.

LNN \cite{RF1-01}, HNN \cite{RF1-07}, and GLNN \cite{RF1-02} remove the distinction between the kinetic and potential energies in favour of directly outputting either the Lagrangian or Hamiltonian. This relaxes the inbuilt assumptions that the kinetic energy is homogeneously quadratic in generalised velocity, and that the potential energy is not a function of the generalised velocity. Hence, these architectures can be applied to relativistic systems and systems that involve charged particles moving through magnetic fields. GLNN also incorporates a friction model that is polynomial in generalised velocity.

Collectively, the current architectures are able to specialise for conservative systems (LNN, HNN, CHNN), incorporate dissipation models (DeLaN 4EC, Dissipative SymODEN, GLNN), and incorporate externally specified generalised forces (DeLaN, DeLaN 4EC, SymODEN, Dissipative SymODEN). However, none of these architectures enable the use of externally specified generalised coordinates.

Externally specified generalised coordinates are a type of system input for representing servomechanisms, stepper motors, current sources, and volumetric pumps. This is in contrast to externally specified generalised forces, which can represent force/torque driven motors and voltage sources. To clarify the significance of this: a voltage source is theoretically interchangeable with a current source, e.g. by calculating the voltage across the current source, but to perform this calculation requires knowing the dynamical model. Thus, the application scenarios for each are distinct \cite{RF1-29,RF1-27,RF1-31}.

This article presents ServoLNN, a new DeLaN-derived architecture to model dynamical systems that incorporate externally specified generalised coordinates; e.g. systems that are driven by servomechanisms. ServoLNN is compatible with the other DeLaN-derived architectures, thus enabling the modelling of systems with mixed types of external interaction.

As an application example to demonstrate the utility provided by incorporating externally specified generalised coordinates into ServoLNN, we apply ServoLNN to model a crane in \secref{sec:evaluation}. By using a servo-driven dynamical model, the crane pendulum dynamics are decoupled from the dynamics of the cart motor and motor driver \cite{BJ3-14}. This decoupling can be justified when high performance and non-backdrivable motors are used \cite{BJ3-14}. Examples from research, where this type of model is applied, are \cite{BJ3-14,BJ4-07,BJ2-43}.

The contributions of this article are the
\begin{itemize}
    \item Description and derivation of ServoLNN.
    \item Derivation of how to use ServoLNN to obtain the system energies, power, rate of work, and the forces that drive the servomechanisms.
    \item Description of the possible family of solutions, and the resolution to obtain a unique solution.
    \item PyTorch implementation \href{https://github.com/Brandon-Johns/ServoLNN}{https://github.com/Brandon-Johns/ServoLNN}.
\end{itemize}

\secref{sec:notation} describes the notation used in this article.
\secref{sec:theory} reforms the relevant physical equations to be compatible with the ServoLNN architecture.
\secref{sec:implementation} describes the architecture and loss function.
\secref{sec:evaluation} evaluates an implementation scenario.
\secref{sec:conclusions} concludes the article.

\section{Notation and matrix derivatives} \label{sec:notation}
The notation in this article writes scalars in non-bold, vectors in bold-lowercase, and matrices in bold-uppercase. Right-superscripts are used only for powers or the matrix transpose. Right-subscripts of $i$, $j$, or $k$ denote the index to a parent vector, matrix, or set. Unless specified otherwise, other right-subscripts and left-scripts are used to distinguish between variables.

Centred dots above variables denote the single, $\dot{a}$, and double, $\ddot{a}$, time derivatives.

Square brackets contextually notate matrix concatenation or the argument of a derivative operator. Where otherwise ambiguous, $a(b)$ notates a function and its arguments. Hence, $a \cdot b$ contextually represents either scalar multiplication or the dot product. In unambiguous cases, $ab$ contextually represents either multiplication by a scalar or the Kronecker product. The Hadamard product is represented by $a \odot b$.

The matrix derivatives $\dot{\mathbf{C}}$ and $\frac{\partial \mathbf{C}}{\partial a}$ are element-wise operations. Likewise, when the matrix is in the denominator. Submatrices of the gradient and Hessian matrices are denoted as

\begin{equation}
    \frac{\partial c}{\partial \mathbf{a}}
    \equiv \begin{bmatrix}
        \frac{\partial c}{\partial a_1} \\
        \vdots  \\
        \frac{\partial c}{\partial a_n} \\
    \end{bmatrix}
    ,\quad
    \frac{\partial^2 c}{\partial \mathbf{a} \partial \mathbf{b}^T}
    \equiv \begin{bmatrix}
        \frac{\partial^2 c}{\partial a_1 \partial b_1} & \cdots & \frac{\partial^2 c}{\partial a_1 \partial b_n} \\
        \vdots & \ddots & \vdots  \\
        \frac{\partial^2 c}{\partial a_m \partial b_1} & \cdots & \frac{\partial^2 c}{\partial a_m \partial b_n}
    \end{bmatrix}
\end{equation}

\noindent
where $n$ and $m$ are the sizes of $\mathbf{b}$ and $\mathbf{a}$, respectively. 

The partial derivative of a matrix with respect to a vector follows the same notation, i.e. $\frac{\partial \mathbf{C}}{\partial \mathbf{a}}$. In instances where this is used, operations on the resultant rank-3 tensor are disambiguated in the text following the introduction of the term.

The following well-known matrix transpose identity is applied many times throughout this article.

\begin{equation} \label{eq:transpose-identity}
    (\mathbf{A}\mathbf{B}\mathbf{C})^\text{T} = \mathbf{C}^\text{T}\mathbf{B}^\text{T}\mathbf{A}^\text{T}
\end{equation}

\section{Lagrangian Mechanics for ServoLNN} \label{sec:theory}
The neural network within ServoLNN represents a function to evaluate the mass matrix and potential energy for a system. The inputs to the neural network are the values of the generalised coordinates, and the outputs are the potential energy and a quantity that is constructed into the mass matrix. These intermediary values are then processed by hard-coded physical equations to produce the final outputs. The final outputs include the system energies, power, rate of work, generalised accelerations, and generalised forces.

In this section, we manipulate the relevant physical equations to be compatible with the ServoLNN architecture. This requires manipulating the equations to be written in terms of the potential energy and mass matrix, and their partial derivatives with respect to the network inputs (the generalised coordinates). Also used are the generalised velocity, generalised acceleration, and generalised force.

\begin{figure}[h]
	\centering
	\centerline{\includegraphics[width=0.5\textwidth,keepaspectratio]{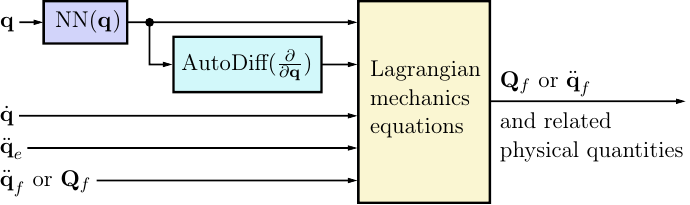}}
	\caption{Diagram of ServoLNN. A more detailed representation is \figref{fig:NN-diagram}. $\text{NN}(\mathbf{q})$ is the neural network, $\mathbf{q}$ are the generalised coordinates, and $\mathbf{Q}$ are the generalised forces.}
	\label{fig:NN-diagram-simple}
\end{figure}

The overall architecture is summarised in \figref{fig:NN-diagram-simple}. This physics-guided architecture and the choice of independent intermediary values intrinsically enforces that all output quantities are consistent with the physical equations, and that any physical relationships between the output quantities are satisfied.

\secref{ssec:EGC} describes how externally specified generalised coordinates can be included in the Euler-Lagrange equations. \secref{ssec:reformed} reforms the equations to be compatible with the neural network. \secref{ssec:quantities} reforms the equations of related physical quantities likewise. \secref{ssec:force} derives an equation to find the generalised forces corresponding to the externally specified generalised coordinates.

\subsection{Externally Specified Generalised Coordinates} \label{ssec:EGC}
The application of Lagrangian mechanics to systems that are driven by servomechanisms is not commonly formalised \cite{RF1-26}. Although, some related tools include variables which are ``prescribed functions of time'' \cite{RFb-05} (also called ``prescribed motions'' \cite{RFb-12}), rheonomic constraints \cite{RFb-07}, moving constraints \cite{RFb-12}, and co-energy \cite{RF1-29,RF1-27}. In this section, we combine and extend these tools into an abstraction that we name externally specified generalised coordinates.

Consider a dynamical system that is driven by the output of a position controlled servomechanism. For example, if the top of a pendulum is constrained by an external mechanism to move according to $\mathbf{p}_{top}=\sin(3t)\hat{\mathbf{e}}_x$. This is a rheonomic constraint; a constraint that explicitly depends on time. Rheonomic constraints are usually non-conservative; they can induce motion in the system. For instance, the prescribed motion of $\mathbf{p}_{top}$ would induce swinging of the bob.

Substituting a rheonomic constraint into the Lagrangian usually results in a time dependent Lagrangian \cite{RF1-24}; the constraint on $\mathbf{p}_{top}$ would be included in the Lagrangian through the kinetic energy term. However, it is undesirable to have explicit time dependence in the neural network, as this would bind the neural network to the specific function of time. For example, to change the constraint to instead enforce $\mathbf{p}_{top}=e^{-5t}\hat{\mathbf{e}}_x$ would require retraining the model.

In the following, we formalise externally specified generalised coordinates as an abstraction to avoid explicit time dependence. Starting from a common statement of the Euler-Lagrange equations \cite{RFb-07}.

\begin{equation} \label{eq:EL-original}
    \frac{d}{dt}\frac{\partial\mathcal{L}}{\partial\dot{q}^{}_{i}}
    - \frac{\partial\mathcal{L}}{\partial q^{}_{i}}
    = Q_i
    ,\quad i\in[1,N_f]
\end{equation}

\noindent
where $\mathcal{L}(\mathbf{q}^{}_{f}, \dot{\mathbf{q}}^{}_{f}, t)$ is the Lagrangian, $t$ is time, $\mathbf{q}^{}_{f}=[q^{}_{1} \cdots q^{}_{N_f}]^{\text{T}}$ is the set of free generalised coordinates, and $\mathbf{Q}_f=[Q_1 \cdots Q_{N_f}]^{\text{T}}$ is the sum of the generalised forces acting on the system, as projected onto configuration space.

The arguments of the Lagrangian can be abstracted to $\mathcal{L}(\mathbf{q}^{}_{f}, \dot{\mathbf{q}}^{}_{f}, \mathbf{u})$, by factoring out expressions of the form $\mathbf{u}(\mathbf{q}^{}_{f}, \dot{\mathbf{q}}^{}_{f}, t)$. If the vector of $\mathbf{u}$ is an argument of the neural network, then its expression can be arbitrarily changed without requirement of retraining. Because the value of $\mathbf{u}$ is only required at the time of solving, this also enables usage in applications where the value is known only just-in-time. For example, if $\mathbf{u}$ are the outputs of servomechanisms that are driven in real-time by a human operator, then the values cannot be known ahead of time, but they can be known just-in-time.

To enable the use of rheonomic constraints while abstracting explicit time dependence, we define the set of $N_e$ number of externally specified generalised coordinates $\mathbf{q}^{}_{e}(t)$. These are substituted into the Lagrangian through $\mathbf{u}(t)=[\mathbf{q}^{}_{e}, \dot{\mathbf{q}}^{}_{e}]$, where the derivatives are required due to the differentiation to obtain kinetic energy. Thus, the arguments of the Lagrangian become $\mathcal{L}(\mathbf{q}^{}_{f}, \dot{\mathbf{q}}^{}_{f}, \mathbf{q}^{}_{e}, \dot{\mathbf{q}}^{}_{e})$. We rewrite this as $\mathcal{L}(\mathbf{q}, \dot{\mathbf{q}})$, where

\begin{equation} \label{eq:q-all}
    \mathbf{q} = \begin{bmatrix} \mathbf{q}^{}_{f} \\ \mathbf{q}^{}_{e} \end{bmatrix},\quad
    \mathbf{q}^{}_{f} = \begin{bmatrix} q^{}_{1} \\ \vdots \\ q^{}_{N_f} \end{bmatrix} ,\quad
    \mathbf{q}^{}_{e} = \begin{bmatrix} q^{}_{N_f+1} \\ \vdots \\ q^{}_{N_f+N_e} \end{bmatrix}
\end{equation}

The physical interpretation of $\mathbf{q}^{}_{e}$ is the same as of the free generalised coordinates. The only difference is that $\mathbf{q}^{}_{e}$ are externally determined and specified, whereas $\mathbf{q}^{}_{f}$ are free variables. This distinction is most important in the forward dynamics problem $\ddot{\mathbf{q}}^{}_f = f(\mathbf{q},\dot{\mathbf{q}},\ddot{\mathbf{q}}^{}_{e}, \mathbf{Q}_f)$, where $\ddot{\mathbf{q}}^{}_{f}$ are dependent variables and $\ddot{\mathbf{q}}^{}_{e}$ are independent variables. For the inverse dynamics problem $\mathbf{Q}_f = f^{-1}(\mathbf{q},\dot{\mathbf{q}},\ddot{\mathbf{q}})$, the distinction is more subtle because both are independent variables.

In practice, externally specified generalised coordinates can be used to represent the outputs of servomechanisms, electrical current sources, volumetric pumps, etc. They are related to flow sources, an abstraction that is used in bond graph modelling \cite{RF1-27}. They are also related to the dual forms of the Lagrangian and Hamiltonian, and the Brayton-Moser equations \cite{RF1-29,RF1-27,RF1-31}. In \secref{ssec:force}, this duality is used to relate the conjugate pairs of the externally specified generalised coordinate and the corresponding generalised force required to cause the prescribed motion.

\subsection{Reformed Euler-Lagrange Equations} \label{ssec:reformed}
This subsection reforms the Euler-Lagrange equations, augmented with externally specified generalised coordinates, to be compatible with the neural network architecture.

The matrix form of the Euler-Lagrange equation is

\begin{equation} \label{eq:EL-vector}
    \frac{d}{dt}\frac{\partial\mathcal{L}}{\partial\dot{\mathbf{q}}^{}_{f}}
    - \frac{\partial\mathcal{L}}{\partial\mathbf{q}^{}_{f}}
    = \mathbf{Q}_f
\end{equation}

Assuming that the kinetic and potential energies, $\mathcal{T}$ and $\mathcal{V}$, are not explicitly time dependent, and that the potential energy is not velocity dependent \cite{RFb-07}, then $\mathcal{L}(\mathbf{q}, \dot{\mathbf{q}})=\mathcal{T}(\mathbf{q}, \dot{\mathbf{q}})-\mathcal{V}(\mathbf{q})$, and \eqref{eq:EL-vector} expands to

\begin{equation} \label{eq:EL-TV}
    \frac{d}{dt}\frac{\partial\mathcal{T}}{\partial\dot{\mathbf{q}}^{}_{f}}
    - \frac{\partial\mathcal{T}}{\partial\mathbf{q}^{}_{f}}
    + \frac{\partial\mathcal{V}}{\partial\mathbf{q}^{}_{f}}
    = \mathbf{Q}_f
\end{equation}

Assuming that the kinetic energy is homogeneously quadratic in generalised velocity \cite{RFb-07}, then it can be written in mass matrix form.

\begin{equation} \label{eq:T-mass-matrix}
    \mathcal{T} = \frac{1}{2} \dot{\mathbf{q}}^{\text{T}} \cdot \mathbf{M} \cdot \dot{\mathbf{q}}
\end{equation}

\noindent
where $\mathbf{M}(\mathbf{q})=\frac{\partial^2 \mathcal{T}}{\partial \dot{\mathbf{q}} \partial \dot{\mathbf{q}}^{\text{T}}}$ is the mass matrix. Partitioning the mass matrix to separate the coefficients of the free and external generalised coordinates gives

\begin{equation}
    \mathbf{M}
    = \begin{bmatrix}
        \mathbf{M}_{ff} & \mathbf{M}_{fe} \\
        \mathbf{M}_{ef} & \mathbf{M}_{ee}
    \end{bmatrix}
\end{equation}

where $\mathbf{M}_{ff}=\frac{\partial^2 \mathcal{T}}{\partial \dot{\mathbf{q}}^{}_{f} \partial \dot{\mathbf{q}}^{\text{T}}_{f}}$, $\mathbf{M}_{ee}=\frac{\partial^2 \mathcal{T}}{\partial \dot{\mathbf{q}}^{}_{e} \partial \dot{\mathbf{q}}^{\text{T}}_{e}}$, $\mathbf{M}_{fe}=\frac{\partial^2 \mathcal{T}}{\partial \dot{\mathbf{q}}^{}_{f} \partial \dot{\mathbf{q}}^{\text{T}}_{e}}$, and $\mathbf{M}_{ef}=\mathbf{M}_{fe}^\text{T}$. The transpose relation is due to the symmetry of the mass matrix. Substituting the partitioned mass matrix into \eqref{eq:T-mass-matrix}, and applying the transpose relation, gives

\begin{equation}
    \mathcal{T}
    = \frac{1}{2} \dot{\mathbf{q}}^{\text{T}}_{f} \cdot \mathbf{M}_{ff} \cdot \dot{\mathbf{q}}^{}_{f}
    + \frac{1}{2} \dot{\mathbf{q}}^{\text{T}}_{e} \cdot \mathbf{M}_{ee} \cdot \dot{\mathbf{q}}^{}_{e}
    + \dot{\mathbf{q}}^{\text{T}}_{f} \cdot \mathbf{M}_{fe} \cdot \dot{\mathbf{q}}^{}_{e}
\end{equation}

Substituting this into the first term of \eqref{eq:EL-TV} and computing the partial derivative then gives

\begin{equation}
    \frac{d}{dt}\frac{\partial\mathcal{T}}{\partial\dot{\mathbf{q}}^{}_{f}}
    = \frac{d}{dt}\left[\mathbf{M}_{ff} \cdot \dot{\mathbf{q}}^{}_{f} + \mathbf{M}_{fe} \cdot \dot{\mathbf{q}}^{}_{e}\right]
\end{equation}

Applying the product rule results in

\begin{equation} \label{eq:EL-MM-term1}
    \frac{d}{dt}\frac{\partial\mathcal{T}}{\partial\dot{\mathbf{q}}^{}_{f}}
    = \mathbf{M}_{ff} \cdot \ddot{\mathbf{q}}^{}_{f} + \mathbf{M}_{fe} \cdot \ddot{\mathbf{q}}^{}_{e}
    + \dot{\mathbf{M}}_{ff} \cdot \dot{\mathbf{q}}^{}_{f} + \dot{\mathbf{M}}_{fe} \cdot \dot{\mathbf{q}}^{}_{e}
\end{equation}

The total derivative of the mass matrix can be found through the chain rule, where the matrix derivative is an element-wise operation. Hence, the derivative of each element is

\begin{equation} \label{eq:dMdt}
    \dot{M}_{ij} = \frac{\partial M_{ij}}{\partial \mathbf{q}^{\text{T}}} \cdot \dot{\mathbf{q}}
\end{equation}

Moving on to the second term of \eqref{eq:EL-TV}, substituting in \eqref{eq:T-mass-matrix} gives 

\begin{align}
    \frac{\partial}{\partial\mathbf{q}^{}_{f}}\left[\frac{1}{2} \dot{\mathbf{q}}^{\text{T}} \cdot \mathbf{M} \cdot \dot{\mathbf{q}}\right]
    =& \sum_{i,j=1}^{N_f+N_e}\left(\frac{1}{2} \dot{q}^{}_{i} \frac{\partial M_{ij}}{\partial\mathbf{q}^{}_{f}} \dot{q}^{}_{j}\right) \label{eq:EL-MM-term2-a}\\
    =& \frac{1}{2} \dot{\mathbf{q}}^{\text{T}} \cdot \frac{\partial \mathbf{M}}{\partial\mathbf{q}^{}_{f}} \cdot \dot{\mathbf{q}} \label{eq:EL-MM-term2-b}
\end{align}

Note that \eqref{eq:EL-MM-term2-b} contains a rank-3 tensor (see \secref{sec:notation}). Resultantly, the dot-product notation is ambiguous. For this case, the meaning of the notation should be interpreted by referring to the unambiguous \eqref{eq:EL-MM-term2-a}.

Substituting \eqref{eq:EL-MM-term1} and \eqref{eq:EL-MM-term2-b} into the Euler-Lagrange equation, \eqref{eq:EL-TV}, finally results in

\begin{equation} \label{eq:EL-MM}
    \mathbf{M}_{ff} \cdot \ddot{\mathbf{q}}^{}_{f} + \mathbf{M}_{fe} \cdot \ddot{\mathbf{q}}^{}_{e}
    + \dot{\mathbf{M}}_{ff} \cdot \dot{\mathbf{q}}^{}_{f} + \dot{\mathbf{M}}_{fe} \cdot \dot{\mathbf{q}}^{}_{e}
    - \frac{1}{2} \dot{\mathbf{q}}^{\text{T}} \cdot \frac{\partial \mathbf{M}}{\partial\mathbf{q}^{}_{f}} \cdot \dot{\mathbf{q}}
    + \frac{\partial\mathcal{V}}{\partial\mathbf{q}^{}_{f}}
    = \mathbf{Q}_f
\end{equation}

Comparing this to DeLaN \cite{RF1-06}, the formulations are related. The DeLaN formulation \cite[Equation~6]{RF1-06} can be seen to be a special case of \eqref{eq:EL-MM} that occurs when $N_e=0$.

\subsection{Equations for Physical Quantities} \label{ssec:quantities}
This section forms expressions for various different physical quantities which can be found using only the terms in \eqref{eq:EL-MM}. Thus allows these quantities to be evaluated concurrency to evaluating the Euler-Lagrange equation.

The inverse dynamics solution is \eqref{eq:EL-MM}. The individual contributions of the inertial, centrifugal, Coriolis, and conservative generalised forces can be separated into

\begin{equation} \label{eq:EL-MM-grouped}
    \mathbf{Q}_{ff,m} + \mathbf{Q}_{fe,m}
    + \mathbf{Q}_{f,c}
    + \mathbf{Q}_{f,g}
    = \mathbf{Q}_f
\end{equation}

\noindent
where the inertial forces are $\mathbf{Q}_{ff,m}=\mathbf{M}_{ff} \cdot \ddot{\mathbf{q}}^{}_{f}$ and $\mathbf{Q}_{fe,m}=\mathbf{M}_{fe} \cdot \ddot{\mathbf{q}}^{}_{e}$, the combined centrifugal and Coriolis forces are

\begin{equation}
    \mathbf{Q}_{f,c}=\dot{\mathbf{M}}_{ff} \cdot \dot{\mathbf{q}}^{}_{f} + \dot{\mathbf{M}}_{fe} \cdot \dot{\mathbf{q}}^{}_{e} - \frac{1}{2} \dot{\mathbf{q}}^{\text{T}} \cdot \frac{\partial \mathbf{M}}{\partial\mathbf{q}^{}_{f}} \cdot \dot{\mathbf{q}}
\end{equation}

\noindent
and $\mathbf{Q}_{f,g}=\frac{\partial\mathcal{V}}{\partial\mathbf{q}^{}_{f}}$ are the conservative forces.

The forward dynamics solution is rearrangement of \eqref{eq:EL-MM} into

\begin{equation} \label{eq:EL-forwardDynamics}
    \ddot{\mathbf{q}}^{}_{f} = \mathbf{M}_{ff}^{-1} \cdot \left( \mathbf{Q}_f - \mathbf{Q}_{fe,m} - \mathbf{Q}_{f,c} - \mathbf{Q}_{f,g} \right)
\end{equation}

The kinetic energy is given by \eqref{eq:T-mass-matrix}, and the potential energy is $\mathcal{V}$. The total system energy is $E=\mathcal{T}+\mathcal{V}$.

The rate of change of kinetic energy can be obtained from applying the product rule to \eqref{eq:T-mass-matrix}.

\begin{equation}
    \dot{\mathcal{T}}
    = \frac{1}{2} \ddot{\mathbf{q}}^\text{T} \cdot \mathbf{M} \cdot \dot{\mathbf{q}}
    + \frac{1}{2} \dot{\mathbf{q}}^{\text{T}} \cdot \dot{\mathbf{M}} \cdot \dot{\mathbf{q}}
    + \frac{1}{2} \dot{\mathbf{q}}^{\text{T}} \cdot \mathbf{M} \cdot \ddot{\mathbf{q}}
\end{equation}

This can be reduced by applying the matrix transpose identity \eqref{eq:transpose-identity} and using the symmetry of the mass matrix, resulting in

\begin{equation} \label{eq:dTdt-defaultForm}
    \dot{\mathcal{T}}
    = \dot{\mathbf{q}}^{\text{T}} \cdot \mathbf{M} \cdot \ddot{\mathbf{q}}
    + \frac{1}{2} \dot{\mathbf{q}}^{\text{T}} \cdot \dot{\mathbf{M}} \cdot \dot{\mathbf{q}}
\end{equation}

The second term of \eqref{eq:dTdt-defaultForm} can be related to the centrifugal and Coriolis forces as follows. First, it is expanded into a summation. Then the total derivative is expanded with the chain rule according to \eqref{eq:dMdt}, and by using the identity \eqref{eq:transpose-identity}. The result is then rearranged.

\begin{align}
    \frac{1}{2} \dot{\mathbf{q}}^{\text{T}} \cdot \dot{\mathbf{M}} \cdot \dot{\mathbf{q}}
    =& \sum_{i,j=1}^{N_f+N_e}\left(\frac{1}{2} \dot{q}^{}_{i} \dot{M}_{ij} \cdot \dot{q}^{}_{j}\right) \nonumber\\
    =& \sum_{i,j=1}^{N_f+N_e}\left(\frac{1}{2} \dot{q}^{}_{i} \left(\dot{\mathbf{q}}^{\text{T}} \cdot \frac{\partial M_{ij}}{\partial \mathbf{q}}\right) \dot{q}^{}_{j}\right) \nonumber\\
    =& \dot{\mathbf{q}}^{\text{T}} \cdot \sum_{i,j=1}^{N_f+N_e}\left(\frac{1}{2} \dot{q}^{}_{i} \frac{\partial M_{ij}}{\partial \mathbf{q}} \dot{q}^{}_{j}\right) \nonumber\\
    =& \dot{\mathbf{q}}^{\text{T}} \cdot \left(\frac{1}{2} \dot{\mathbf{q}}^{\text{T}} \frac{\partial \mathbf{M}}{\partial \mathbf{q}} \dot{\mathbf{q}}\right)
\end{align}

Adding/subtracting the left/right hand sides of this to \eqref{eq:dTdt-defaultForm} results in the form

\begin{equation} \label{eq:dTdt}
    \dot{\mathcal{T}}
    = \dot{\mathbf{q}}^{\text{T}} \cdot \left(\mathbf{M} \cdot \ddot{\mathbf{q}}
    + \dot{\mathbf{M}} \cdot \dot{\mathbf{q}}
    - \frac{1}{2} \dot{\mathbf{q}}^{\text{T}} \frac{\partial \mathbf{M}}{\partial \mathbf{q}} \dot{\mathbf{q}}\right)
\end{equation}
 
The interpretation and significance of this result is discussed in \secref{ssec:force}.

The rate of change of potential energy can be obtained from the chain rule.

\begin{equation} \label{eq:dVdt}
    \dot{\mathcal{V}} = \frac{\partial \mathcal{V}}{\partial \mathbf{q}^{\text{T}}} \cdot \dot{\mathbf{q}}
\end{equation}

Finally, the rate of change of total system energy is

\begin{equation} \label{eq:dEdt}
    \dot{E}=\dot{\mathcal{T}}+\dot{\mathcal{V}}
\end{equation}

These equations can be seen to be general cases of the corresponding DeLaN formulations in \cite{RF1-06,RF1-03}, where the DeLaN formulations occur when $N_e=0$.

\subsection{The Equivalent Force} \label{ssec:force}
A quantity that is unique to our formulation with externally specified generalised coordinates is the corresponding generalised force $\mathbf{Q}_e$. That is to say, if the externally specified generalised coordinates were to be converted to free generalised coordinates with accompanying externally specified generalised forces, then $\mathbf{Q}_e$ would be the forces that result in the same motion (this concept is also described in \cite{RFb-12} as the ``driving forces required to establish known motions''). In terms of servomechanisms, $\mathbf{Q}_e$ is the force that drives the servomechanism. This section solves for $\mathbf{Q}_e$.

The rate of work done on the system by the generalised forces can be expressed as force multiplied by velocity, as performed element-wise.

\begin{equation} \label{eq:dWdt}
    \dot{\mathbf{W}} = \dot{\mathbf{q}} \odot \mathbf{Q}
\end{equation}

\noindent
and summing over each degree of freedom gives $\dot{W}_{\text{total}} = \sum_{i}(\dot{W_i}) = \dot{\mathbf{q}}^{\text{T}} \cdot \mathbf{Q}$, where

\begin{equation}
    \mathbf{Q} = \begin{bmatrix} \mathbf{Q}_f \\ \mathbf{Q}_e \end{bmatrix}
\end{equation}

According to the first law of thermodynamics, the rate of change of total system energy is equal to the rate of work done on the system, $\dot{E}=\dot{W}_{\text{total}}$. Therefore, substituting \eqref{eq:dTdt} and \eqref{eq:dVdt} into $\dot{E}$, and equating to $\dot{W}_{\text{total}}$, gives

\begin{equation} \label{eq:powerConservation}
    \dot{\mathbf{q}}^{\text{T}} \cdot \mathbf{Q}
    = \dot{\mathbf{q}}^{\text{T}} \cdot \left(\mathbf{M} \cdot \ddot{\mathbf{q}}
    + \dot{\mathbf{M}} \cdot \dot{\mathbf{q}}
    - \frac{1}{2} \dot{\mathbf{q}}^{\text{T}} \frac{\partial \mathbf{M}}{\partial \mathbf{q}} \dot{\mathbf{q}}
    + \frac{\partial \mathcal{V}}{\partial \mathbf{q}}\right)
\end{equation}

This is closely related to the Euler-Lagrange equation \eqref{eq:EL-MM}. Let $\mathbf{Q}'$ be the solution to the expression inside of the brackets.

\begin{equation} \label{eq:EL-MM-extended}
    \mathbf{Q}'
    = \mathbf{M} \cdot \ddot{\mathbf{q}}
    + \dot{\mathbf{M}} \cdot \dot{\mathbf{q}}
    - \frac{1}{2} \dot{\mathbf{q}}^{\text{T}} \frac{\partial \mathbf{M}}{\partial \mathbf{q}} \dot{\mathbf{q}}
    + \frac{\partial \mathcal{V}}{\partial \mathbf{q}}
\end{equation}

Partitioning this expression into $\mathbf{Q}'_f$ and $\mathbf{Q}'_e$, and comparing to \eqref{eq:EL-MM}, shows that $\mathbf{Q}'_f\equiv\mathbf{Q}_f$. However, this does not necessarily imply that $\mathbf{Q}'_e$ is equivalent to $\mathbf{Q}_e$.

Nevertheless, we can deduce that they are equivalent by considering the following. If all of $\mathbf{q}^{}_{e}$ were considered to be free generalised coordinates instead of externally specified generalised coordinates, then they would be included in the set of $\mathbf{q}^{}_{f}$, and resultantly, \eqref{eq:EL-MM-extended} and \eqref{eq:EL-MM} would be the same expression. Therefore, $\mathbf{Q}'_e\equiv\mathbf{Q}_e$. The partitioned solution for $\mathbf{Q}_e$ is

\begin{equation} \label{eq:EL-MM-external}
    \mathbf{Q}_e
    = \mathbf{M}_{ee} \cdot \ddot{\mathbf{q}}^{}_{e} + \mathbf{M}_{ef} \cdot \ddot{\mathbf{q}}^{}_{f}
    + \dot{\mathbf{M}}_{ee} \cdot \dot{\mathbf{q}}^{}_{e} + \dot{\mathbf{M}}_{ef} \cdot \dot{\mathbf{q}}^{}_{f}
    - \frac{1}{2} \dot{\mathbf{q}}^{\text{T}} \cdot \frac{\partial \mathbf{M}}{\partial\mathbf{q}^{}_{e}} \cdot \dot{\mathbf{q}}
    + \frac{\partial\mathcal{V}}{\partial\mathbf{q}^{}_{e}}
\end{equation}

Furthermore, this can be separated into the individual contributions of the inertial, centrifugal, Coriolis, and conservative generalised forces, in the same way as \eqref{eq:EL-MM-grouped}.

\section{Neural Network implementation} \label{sec:implementation}
\subsection{Neural Network} \label{ssec:nn}
The neural network part of our architecture derives from the original DeLaN architecture \cite{RF1-04}. A core idea of the DeLaN-derived architectures is that the neural network itself is treated as a mathematical function. Because our architecture is multiheaded, it is represented by a collection of functions.

\begin{equation}
    \mathcal{V}(\mathbf{q})
    , \quad
    \mathbf{l}_{lower}(\mathbf{q})
    , \quad
    \mathbf{l}_{diag}(\mathbf{q})
\end{equation}

Internally, these functions are a neural network with alternating linear layers and activation functions.

All three functions share the same inputs, $\mathbf{q}$. The potential energy network head is a scalar-valued function. It is directly substituted into the previously derived equations. The other two network heads are vector-valued functions. They are jointly assembled into a single lower-triangular matrix.

\begin{equation}
    \mathbf{L}(\mathbf{q})
    = \begin{bmatrix}
        l_{diag,1} & & & 0 \\
        l_{lower,1} & l_{diag,2} & & \\
        l_{lower,2} & l_{lower,3} & l_{diag,3} & \\
        \vdots & \vdots & \ddots & \ddots \\
    \end{bmatrix}
\end{equation}

where the specific ordering of elements does not matter. It is only important that $\mathbf{l}_{diag}$ are placed on the diagonal, and the order does not change between network evaluations. The mass matrix is then constructed by reversing the Cholesky decomposition.

\begin{equation} \label{eq:cholesky-with-epsilon}
    \mathbf{M}(\mathbf{q}) = \mathbf{L}\mathbf{L}^{\text{T}} + \epsilon\mathbf{I}
\end{equation}

\noindent
where $\mathbf{I}$ is an identity matrix, and $\epsilon$ is a user-defined constant positive scalar. The value of $\mathbf{l}_{diag}$ is constrained to be greater than or equal to zero through the use of a ReLU output layer. In combination of the Cholesky decomposition and the addition of $\epsilon$ to the diagonal, this intrinsically causes the mass matrix to be well-conditioned and positive-definite. This aspect of the design is an implementation of physics-guided architecture.

Likewise, the choice of network inputs enforces that $\mathcal{V}$ is not a function of generalised velocity, and that $\mathcal{T}$ is homogeneously quadratic in generalised velocity.

To evaluate the reformed equations presented in \secref{sec:theory}, $\mathbf{M}$ and $\mathcal{V}$ can be substituted directly in. Their partial derivatives with respect to the inputs $\mathbf{q}$ can be computed through automatic differentiation.

Additionally required to evaluate the reformed equations are the values of $\dot{\mathbf{q}}$ and $\ddot{\mathbf{q}}^{}_{e}$, and either $\ddot{\mathbf{q}}^{}_{f}$ or $\mathbf{Q}_f$, depending on if the forward or inverse dynamics calculation is required. These values do not pass through the network layers; they only operate on the output.

\begin{figure}[h]
	\centering
	\centerline{\includegraphics[width=\textwidth,keepaspectratio]{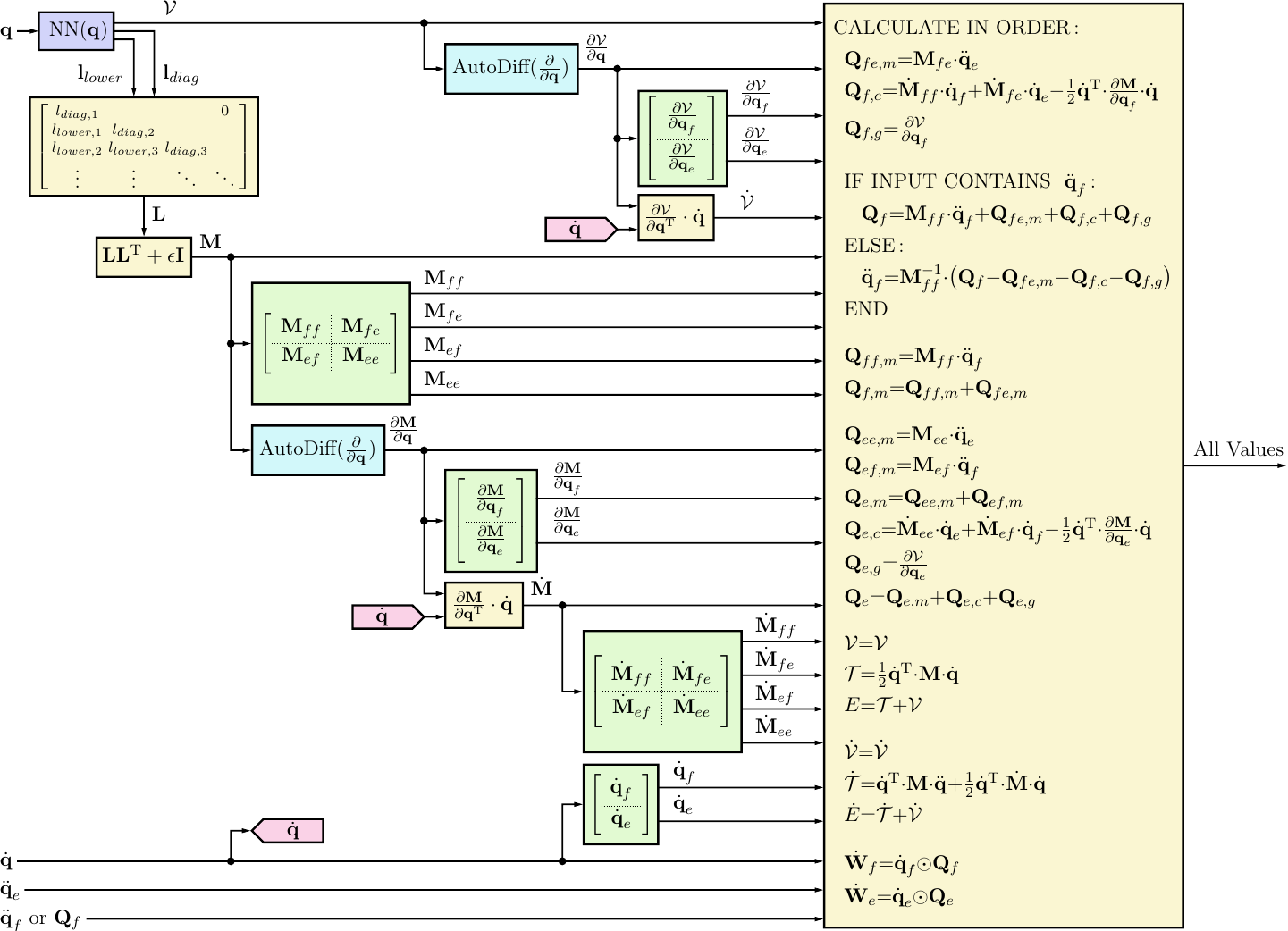}}
	\caption{Diagram of ServoLNN. The depicted structure, as well as the mathematical symbols and equations, are introduced in \secref{sec:theory} and \secref{ssec:nn}. $\text{NN}(\mathbf{q})$ represents the multiheaded neural network.}
	\label{fig:NN-diagram}
\end{figure}

\figref{fig:NN-diagram} details the overall ServoLNN architecture, as described in this section and \secref{sec:theory}. The order of calculations is depicted, and an `if' statement shows how the architecture can predict either forward or inverse dynamics, depending on the provided input.

\subsection{Loss function} \label{ssec:loss}
The loss function is the sum of the losses for inverse dynamics, forward dynamics, and power.

\begin{equation}
    \mathbb{L} = \mathbb{L}_{\text{inv}} + \mathbb{L}_{\text{fwd}} + \mathbb{L}_{\text{power}}
\end{equation}

Let the estimated inverse dynamics solution, as calculated by substituting the outputs of the neural network into \eqref{eq:EL-MM}, be $\hat{\mathbf{Q}}_f(\mathbf{q},\dot{\mathbf{q}},\ddot{\mathbf{q}})$. Likewise, let the estimated forward dynamics solution from evaluating \eqref{eq:EL-forwardDynamics} be $\hat{\ddot{\mathbf{q}}}_f(\mathbf{q},\dot{\mathbf{q}},\ddot{\mathbf{q}}^{}_{e}, \mathbf{Q}_f)$. The inverse dynamics and forward dynamics losses are

\begin{equation} \label{eq:invLoss}
    \mathbb{L}_{\text{inv}} = \left\lVert \hat{\mathbf{Q}}_f - \mathbf{Q}_f \right\rVert^2
\end{equation}

\begin{equation} \label{eq:fwdLoss}
    \mathbb{L}_{\text{fwd}} = \left\lVert \hat{\ddot{\mathbf{q}}}_f - \ddot{\mathbf{q}}^{}_{f} \right\rVert^2
\end{equation}

One of, but not both of $\mathbb{L}_{\text{inv}}$ and $\mathbb{L}_{\text{fwd}}$ may optionally be omitted from the loss function.

The expressions for forward and inverse dynamics, and hence the loss functions $\mathbb{L}_{\text{inv}}$ and $\mathbb{L}_{\text{fwd}}$, do not capture the terms in $\mathcal{V}$ and $\mathbf{M}_{ee}$ that depend solely on $\mathbf{q}^{}_{e}$ (that are independent of $\mathbf{q}^{}_{f}$). Hence, the network can be trained to predict the correct forward and inverse dynamics by using $\mathbb{L}_{\text{inv}}$ or $\mathbb{L}_{\text{fwd}}$ alone, but the predicted energy, power, and force decompositions would not be as expected. The network would converge on one valid solution within the family of infinitely many possible solutions.

This can be resolved only if the training data contains the generalised forces $\mathbf{Q}_e$ that correspond to the externally specified generalised coordinates. In this case, then the problem can be resolved by additionally using \eqref{eq:EL-MM-external} to calculate the forward and inverse dynamics losses of $\mathbf{Q}_e$.

The power loss function is

\begin{equation} \label{eq:powerLoss}
    \mathbb{L}_{\text{power}} = \left\lVert \hat{\dot{E}} - \dot{\mathbf{q}}^{\text{T}} \cdot \mathbf{Q} \right\rVert^2
\end{equation}

\noindent
where $\hat{\dot{E}}(\mathbf{q},\dot{\mathbf{q}},\ddot{\mathbf{q}})$ is the estimated rate of change of total system energy from applying \eqref{eq:dEdt}. The power loss is the difference between this, and the rate of work done on the system.

The power loss fully captures $\mathcal{V}$ and the mass matrix. However, calculating the power loss requires that the training data contains $\mathbf{Q}_e$. Alternatively, if these are not known, then they can be substituted for the estimations $\hat{\mathbf{Q}}_e(\mathbf{q},\dot{\mathbf{q}},\ddot{\mathbf{q}})$ obtained from evaluating \eqref{eq:EL-MM-external}. The effect of making this substitution is derived as follows.

Starting from the true power loss \eqref{eq:powerLoss}, replacing $\mathbf{Q}_e$ with $\hat{\mathbf{Q}}_e$ results in the estimated power loss.

\begin{equation} \label{eq:powerLossEstimated}
    \hat{\mathbb{L}}_{\text{power}} = \left\lVert \hat{\dot{E}} - \dot{\mathbf{q}}^{\text{T}}_{f} \cdot \mathbf{Q}_f - \dot{\mathbf{q}}^{\text{T}}_{e} \cdot \hat{\mathbf{Q}}_e \right\rVert^2
\end{equation}

Expanding the terms with \eqref{eq:dEdt}, \eqref{eq:dVdt}, \eqref{eq:dTdt}, and \eqref{eq:EL-MM-external}, and then cancelling and regrouping the terms reduces this to

\begin{equation}
    \hat{\mathbb{L}}_{\text{power}} = \left\lVert \dot{\mathbf{q}}^{\text{T}} \cdot \hat{\mathbf{Q}}_f - \dot{\mathbf{q}}^{\text{T}}_{f} \cdot \mathbf{Q}_f \right\rVert^2
\end{equation}

This does not capture the terms in $\mathcal{V}$ and $\mathbf{M}_{ee}$ that depend solely on $\mathbf{q}^{}_{e}$. Nevertheless, the estimated power loss can still be used to stabilise the training in cases where only the forward and inverse dynamic solutions are required. Otherwise, the true power loss \eqref{eq:powerLoss} should be used.

For a physical interpretation of why the solution becomes a family of solutions, consider a position driven pendulum cart; a model that is commonly used in research to control cranes \cite{BJ3-14,BJ4-07,BJ2-43}. The mass of the cart would appear in the $\mathbf{M}_{ee}$ term, and also in $\mathcal{V}$ if the cart travels up a slope. Changing the mass of the cart changes the equivalent driving force $\mathbf{Q}_e$, as well as the kinetic and potential energies of the cart. However, because position of the cart has been externally specified, the pendulum dynamics are independent of the mass of the cart. Hence, any mass of cart will result in the same solution trajectory.

The general implication is that a given solution trajectory can be associated with multiple very different dynamical systems and system equations \cite{RF1-25}. In application, it may not be important for the network internals to match the correct system. In this case, it would be reasonable to use the estimated power loss instead of the true power loss.

\section{Evaluation} \label{sec:evaluation}
\subsection{Dataset}
The evaluation dataset models a position driven pendulum cart. This model was chosen due to its common application in the control of cranes.

The dataset was generated with Crane Dynamics Simulator \cite{BJ-04-code}, which was first presented in \cite{BJ-04}. The MATLAB ode45 solver was used with relative and absolute tolerances of $10^{-10}$. The data was sampled at the steps that were automatically chosen by the adaptive variable-step-size ODE solver.

\begin{figure}[h]
	\centering
	\centerline{\includegraphics[width=0.4\textwidth,keepaspectratio]{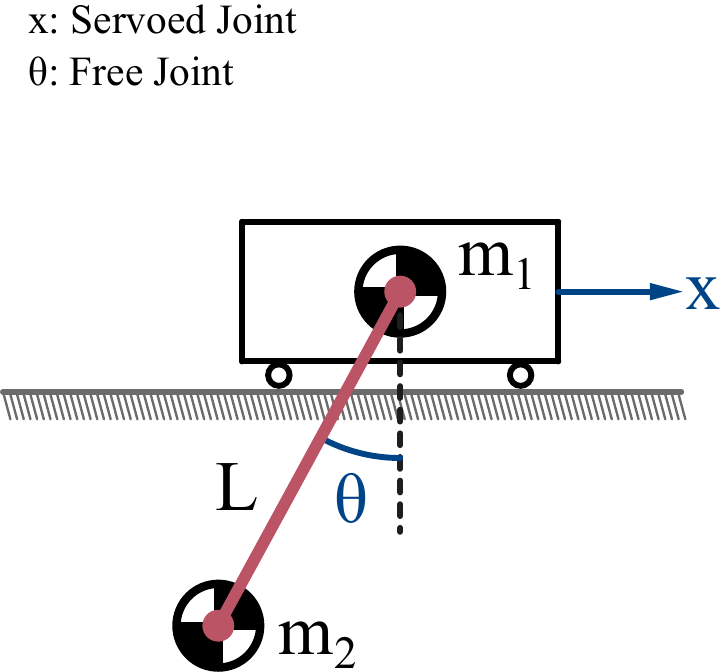}}
	\caption{Diagram of the tested dynamical system.}
	\label{fig:pc-diagram}
\end{figure}

The system is shown in \figref{fig:pc-diagram}. The system parameters were chosen pseudo-randomly as to prevent calculated values from trivially cancelling, as that would risk giving false impression of the performance. Specifically, the values $m_1=0.45$~kg, $m_2=0.13$~kg, $L=1.5$~m, and $g=9.8$~ms\textsuperscript{-2} respectively represent the mass of the cart, the mass of the pendulum bob, the length of the string, and the gravitational acceleration. The free and externally specified generalised coordinates are the angle of the pendulum $\mathbf{q}^{}_{f}=\theta$, and the position of the cart $\mathbf{q}^{}_{e}=x$. The initial conditions for all trials set $x(0)=0$~m and either $\theta(0)=0$ or $\theta(0)=1$~radians. Each trial was $20$~seconds in duration.

\begin{figure}[h]
	\centering
	\centerline{\includegraphics[width=0.5\textwidth,keepaspectratio]{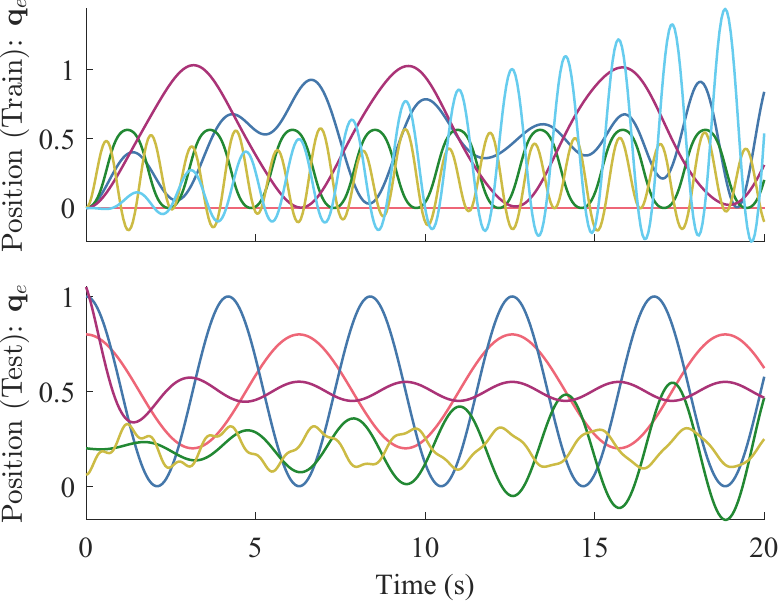}}
	\caption{Trajectories of the externally specified generalised coordinate used during training and testing. Each curve represents a different trial. Units in meters.}
	\label{fig:pc-inputs}
\end{figure}

Data was generated from 11 different trials, each using a different function for $x(t)$. These consisted of cosine waves; some with multiple frequencies, and some with decaying amplitude. One trial held the cart stationary at $x(t)=0$. To increase the variety of data without having to manually create complicated expressions for $x(t)$, 4 of the trials were force driven instead of position driven. The relation between the force and position driven systems, as discussed in \secref{ssec:force}, allowed this data to be replayed into as though it had instead received position input. All inputs are plotted in \figref{fig:pc-inputs}.

\subsection{Training}
The structure of the network follows: As a 2 degrees of freedom system, the network dimensions were $2 \times 64 \times [1,1,2]$; the input size was 2, the single hidden layer had a size of 64, and the sizes of the network heads were 1, 1, and 2. The input and hidden layers were all shared. The value of $\epsilon$ from \eqref{eq:cholesky-with-epsilon} was $0.01$. The activation function was implemented as SoftPlus. These settings were chosen to be similar to the settings used in \cite{RF1-06}.

The training settings follows: The ADAM optimiser was used with a learning rate of $10^{-4}$ and a weight decay of  $10^{-5}$, as chosen to match \cite{RF1-06}. The data batch size was $2048$, as limited by the NVIDIA GeForce GTX 960 GPU used for training. The network was trained until convergence at $10000$ epochs.

It is important to train the system on multiple trajectories to ensure that the network converges on the correct system equations, as given the implications of \cite{RF1-25}, that a given solution trajectory can be associated with multiple very different system equations. Therefore, the training dataset used $4096$ randomly selected samples from $6$ trials. The remaining $5$ trials ($5864$ samples) were used for testing.

\subsection{Evaluation}
The network was trained in two different ways. First, for the case when the ground truth of the equivalent driving force $\mathbf{Q}_e$ is known during training. Then, for the case when it is not known. In each case, the loss function was composed of the sum of all applicable loss functions from \secref{ssec:loss}. In both cases, $\mathbf{Q}_e$ was not known during testing. Both cases used the same network initialisation.

For both cases, the network was trained multiple times over while changing only the random seed between runs. This was done to create statistical measures about the family of solutions. Because not all cases sufficiently converged within the fixed number of epochs (though the trend showed that they would converge if trained for longer), only the models that achieved a total loss below $10^{-2}$ are plotted. This includes 10 models for the case of training with $\mathbf{Q}_e$, and 8 models for the case of training without $\mathbf{Q}_e$.

\begin{figure}[h]
	\centering
	\centerline{\includegraphics[width=0.5\textwidth,keepaspectratio]{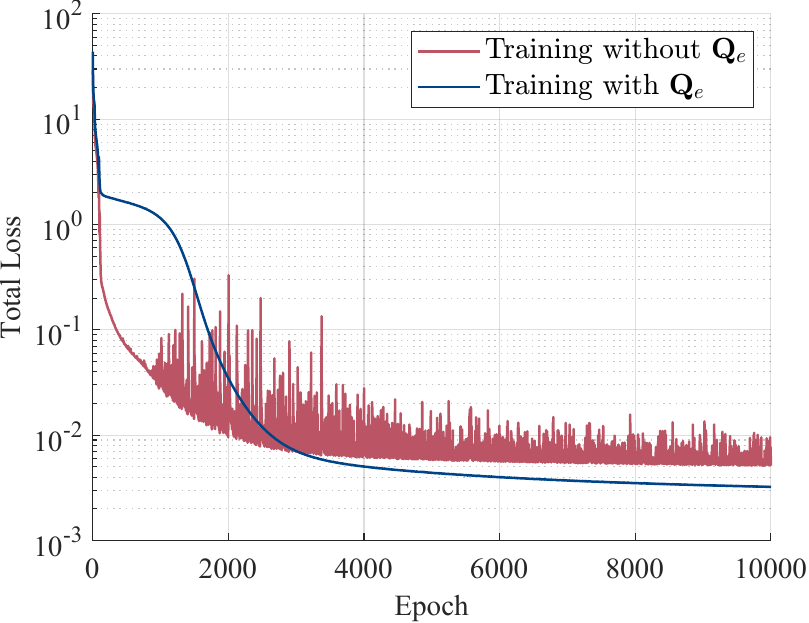}}
	\caption{Total loss during training for each method. Both cases used the same random seed of 42.}
	\label{fig:pc-training}
\end{figure}

The training loss is plotted in \figref{fig:pc-training}. Both cases show a trend of convergence. The non-smoothness of the trend from training without $\mathbf{Q}_e$ is explainable by the existence of a large family of solutions, which requires the optimiser to explore the null space of the loss function. Although the loss function allows a family of solutions, the relatively weak weight decay causes eventual convergence towards a specific solution.

\begin{figure}[htb]
	\centering
	\centerline{\includegraphics[width=1\textwidth,keepaspectratio]{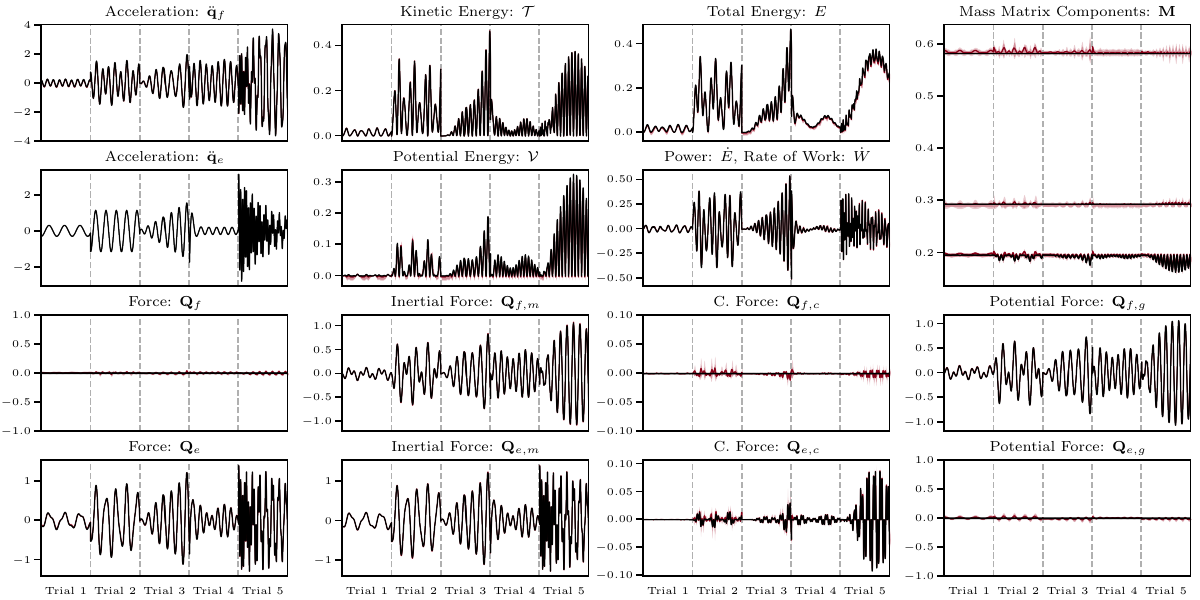}}
	\caption{Results of running the test dataset for the case of training when $\mathbf{Q}_e$ \textbf{is known}. Units in SI. Key: \ColouredCircleK{}~truth. \ColouredCircleD{}~prediction with random seed of 42. \ColouredCircleM{}~mean($\cdot$)$\pm1\sigma$. \ColouredCircleL{}~[min($\cdot$),max($\cdot$)].}
	\label{fig:pc-results-withQe}
\end{figure}

\begin{figure}[htb]
	\centering
	\centerline{\includegraphics[width=1\textwidth,keepaspectratio]{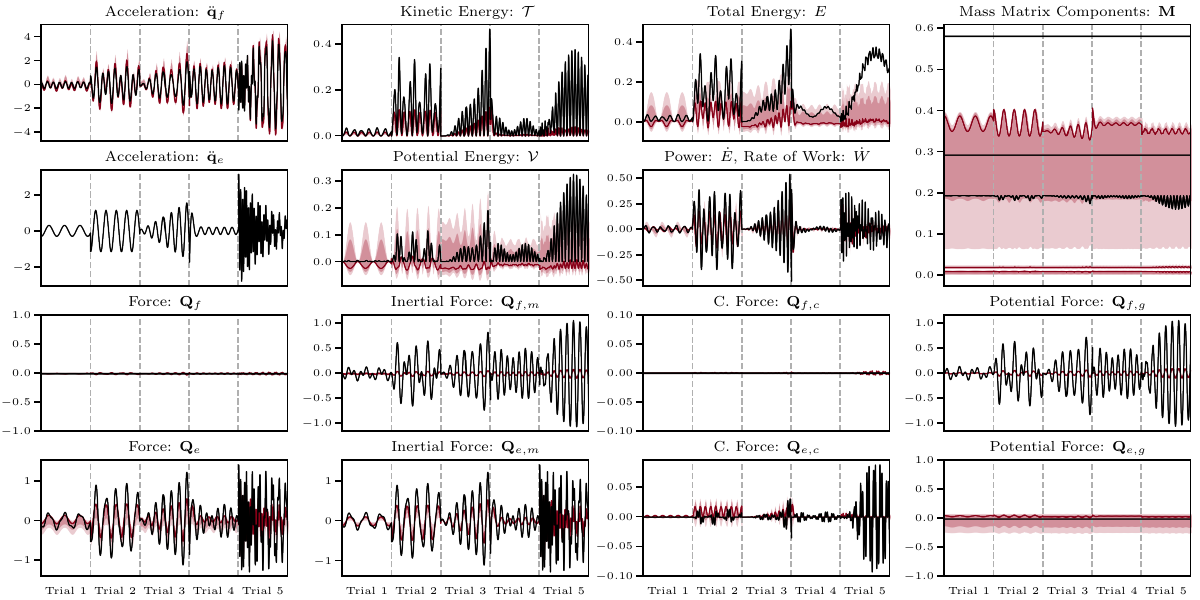}}
	\caption{Results of running the test dataset for the case of training when $\mathbf{Q}_e$ \textbf{is not known}. Units in SI. Key: \ColouredCircleK{}~truth. \ColouredCircleD{}~prediction with random seed of 42. \ColouredCircleM{}~mean($\cdot$)$\pm1\sigma$. \ColouredCircleL{}~[min($\cdot$),max($\cdot$)].}
	\label{fig:pc-results-noQe}
\end{figure}

The results of testing on the model that was trained with $\mathbf{Q}_e$ is \figref{fig:pc-results-withQe}, and the results of testing on the model that was trained without $\mathbf{Q}_e$ is \figref{fig:pc-results-noQe}. To test the null hypothesis, the results of testing on a completely untrained model is \figref{fig:pc-results-notTrained}. Both the forward and inverse dynamics predictions were evaluated, as well as all of the quantities derived in \secref{ssec:quantities} and \secref{ssec:force}. The elements of the mass matrix are also plotted. The externally specified acceleration does not have any associated prediction because it is always specified, hence, in the plots of $\ddot{\mathbf{q}}^{}_{e}$, only the ground truth is shown.

In the plots \figref{fig:pc-results-withQe}, \figref{fig:pc-results-noQe}, and \figref{fig:pc-results-notTrained}, the black lines are the ground truth, and the dark coloured lines are the predictions from when a random seed of 42 is used. The medium and light colour shaded areas are statistical measures from training while changing only the random seed between runs. The medium colour bounds one standard deviation above and below the mean, and the light colour bounds the full range of predictions.

As expected from the analysis in \secref{ssec:loss}, all values converge on the true result when $\mathbf{Q}_e$ is known, and when it is not known, only the forward and inverse dynamics of the free generalised coordinates converge on the true result. Evidently, it is ideal to train with $\mathbf{Q}_e$, when it is available.

While some of the accuracy of the results can be attributed to training, it is important to consider that the derived quantities have been multiplied by known input values (see \figref{fig:NN-diagram}). Statistically, multiplying random predictions with true data causes a relationship between the resultant trend and the true trend. This is shown in the results from testing on the completely untrained network (\figref{fig:pc-results-notTrained}). A less biased interpretation can be obtained by inspecting the non-derived quantities, $\mathcal{V}$, $\mathbf{M}$, and their partial (not total) derivatives.

In \figref{fig:pc-results-withQe}, the mass matrix, potential energy, and potential force plots all closely resemble the ground truth, thereby confirming the successful training and successful learning of the system dynamics.

In \figref{fig:pc-results-noQe}, the success of the training is less evident. The existence of a family of solutions allows error, offset, and scaling of internal values. The only constraint on the internal values is that they must combine to correctly predict the directly controlled quantities, $\ddot{\mathbf{q}}^{}_{f}$ and $\mathbf{Q}_f$. Still, evidence of successful training is shown by the decrease of the loss, as well as by the correct frequencies being contained in $\mathbf{Q}_{f,g}$ and $\mathcal{V}$. This is in contrast to the same values as predicted from testing on the completely untrained network (\figref{fig:pc-results-notTrained}), which do not resemble the data. Regarding the curve $\mathbf{Q}_{e,g}$, which was not captured by the loss function, the shape is an artefact from the random network initialisation; it changes when the random seed is changed.

The inference speed was measured by running the network on CPU, one sample at a time, for the entire test dataset. This emulates usage of the network inside of a real-time control loop, where new sensor samples are received one at a time. In each loop, all force, energy, and power outputs were calculated. An Intel Xeon E3-1231V3 CPU (manufactured in 2014) was used. The measured speed averaged 1.7~ms, 590~Hz.

\FloatBarrier
\section{Conclusions and Future Work} \label{sec:conclusions}
This article presents ServoLNN, a new neural network architecture for modelling dynamical systems that are driven by servomechanisms. Externally specified generalised coordinates are used as an abstraction to the otherwise explicit time dependence. This allows the motion of the servomechanism to be arbitrarily changed without requirement of retraining. The abstraction, combined with the 1.7~ms inference speed enables usage in real-time applications, where the motion is known only just-in-time.

ServoLNN is classed as a DeLaN-derived \cite{RF1-06} neural network architecture. It is compatible with the other DeLaN-derived architectures, enabling the modelling of systems with mixed types of external interaction. The value of extending DeLaN with externally specified generalised coordinates is that it enables servomechanisms to be used as system inputs (e.g. servomotors, stepper motors, current sources). In research to control cranes, this is commonly used to decouple the crane pendulum dynamics from the dynamics of the cart motor and motor driver \cite{BJ3-14}.

This article includes the derivations for obtaining the forward dynamics, inverse dynamics, force decomposition, energies, power, rate of work, and the forces that drive the servomechanisms. A PyTorch implementation of ServoLNN is also provided.

The occurrence of a possible family of solutions, which is resultant of the use of externally specified generalised coordinates, is discussed. The effect of this on the predicted physical quantities is explored, as is the resolution to reduce the family of solutions to a single solution.

Some open challenges for the DeLaN-derived architectures are suggested by \cite{RF1-06}. We additionally suggest further exploration of the implications of \cite{RF1-25} for the additional physical quantities predicted by the DeLaN-derived architectures. Given that some solution trajectories can be associated with multiple very different dynamical systems, it is important to be able to categorise which systems or types of training data are susceptible to this. For these systems, the predicted physical quantities may not be correct. Autonomous systems are expected to be the most susceptible due to only being able to vary the initial conditions when collecting training data.

\section*{CRediT Authorship Contribution Statement}
\textbf{Brandon Johns:} Conceptualization, Data curation, Formal analysis, Investigation, Methodology, Project administration, Resources, Software, Validation, Visualization, Writing - original draft, Writing - review \& editing. 

\textbf{Zhuomin Zhou:} Conceptualization, Investigation, Methodology, Software, Writing - review \& editing.

\textbf{Elahe Abdi:} Supervision, Writing - review \& editing. 

\section*{Data and Code Availability}
The code and data is available at \href{https://github.com/Brandon-Johns/ServoLNN}{https://github.com/Brandon-Johns/ServoLNN}.

\FloatBarrier
\appendix

\section{Additional Plot}
\begin{figure}[h]
	\centering
	\centerline{\includegraphics[width=1\textwidth,keepaspectratio]{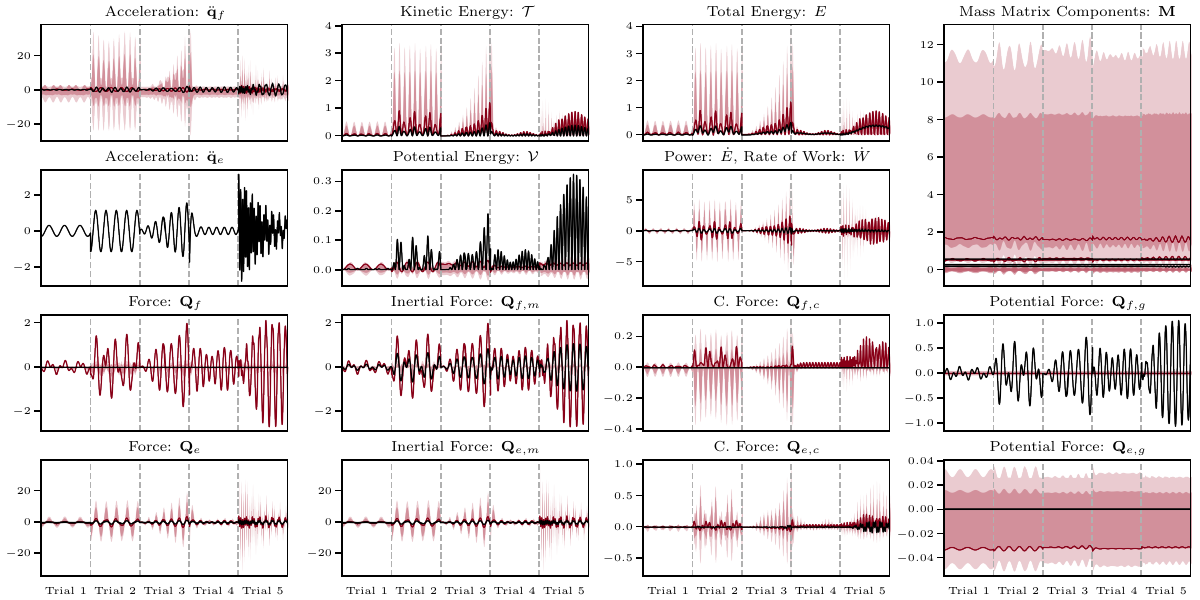}}
	\caption{Results of running the test dataset for the case when \textbf{the network is not trained at all}. Units in SI. Key: \ColouredCircleK{}~truth. \ColouredCircleD{}~prediction with random seed of 42. \ColouredCircleM{}~mean($\cdot$)$\pm1\sigma$. \ColouredCircleL{}~[min($\cdot$),max($\cdot$)].}
	\label{fig:pc-results-notTrained}
\end{figure}

\FloatBarrier
\bibliographystyle{unsrturl}
\bibliography{main.bib}

\end{document}